\documentclass{article}

     \PassOptionsToPackage{numbers, compress}{natbib}

\usepackage[preprint]{neurips_2021}

\usepackage[utf8]{inputenc} 
\usepackage[T1]{fontenc}    
\usepackage{hyperref}       
\usepackage{url}            
\usepackage{booktabs}       
\usepackage{amsfonts}       
\usepackage{nicefrac}       
\usepackage{microtype}      
\usepackage{amsmath}
\usepackage{graphicx}
\usepackage{mathtools}
\usepackage{bm}
\usepackage{multirow}
\usepackage{caption}
\usepackage{subcaption}
\usepackage{alphalph}
\usepackage[dvipsnames]{xcolor}

\usepackage[multiple]{footmisc}
\usepackage{bbm}
\usepackage{lipsum}

\newtheorem{theorem}{Theorem}

\newtheorem{definition}{Definition}

\newtheorem{remark}{Remark}
\newcommand{\R}{\mathbb{R}}

\newcommand{\bx}{\boldsymbol{x}}

\newcommand{\bp}{\boldsymbol{p}}
\newcommand{\by}{\boldsymbol{y}}
\newcommand{\bq}{\boldsymbol{q}}

\title{SympGNNs: Symplectic graph neural networks for identifying high-dimensional Hamiltonian systems and node classification}

\author{%
  \And
  Alan John Varghese\thanks{These authors contributed equally to this work}\\
  School of Engineering\\
  Brown University\\
  Providence, RI 02906 \\
  \texttt{alan\_john\_varghese@brown.edu}\\
  \And
  Zhen Zhang$^*$ \\
  Division of Applied Mathematics\\
  Brown University\\
  Providence, RI 02906 \\
  \texttt{zhen\_zhang1@alumni.brown.edu}\\
  \And
  George Em Karniadakis\thanks{Corresponding author}\\
  Division of Applied Mathematics \\
  Brown University \\
  Providence, RI 02906\\
  \texttt{george\_karniadakis@brown.edu}
}

\begin{document}

\maketitle

\begin{abstract}
Existing neural network models to learn Hamiltonian systems, such as SympNets, although accurate in low-dimensions, struggle to learn the correct  dynamics for high-dimensional many-body systems. 
Herein, we introduce Symplectic Graph Neural Networks (SympGNNs) that can effectively handle system identification in high-dimensional Hamiltonian systems, as well as node classification. SympGNNs combines  symplectic maps with permutation equivariance, a property of graph neural networks. Specifically, we propose two variants of SympGNNs: i) G-SympGNN and ii) LA-SympGNN, arising from different parameterizations of the kinetic and potential energy. We demonstrate the capabilities of SympGNN on two physical examples: a 40-particle coupled Harmonic oscillator, and a 2000-particle molecular dynamics simulation in a two-dimensional Lennard-Jones potential. Furthermore, we demonstrate the performance of SympGNN in the node classification task, achieving accuracy comparable to the state-of-the-art. We also empirically show that SympGNN can overcome the oversmoothing and heterophily problems, two key challenges in the field of graph neural networks.
\end{abstract}

\section{Introduction}
Neural networks perform well in data-rich environments such as computer vision \cite{russakovsky2015imagenet} and natural language processing \cite{achiam2023gpt}, but struggle when the data is scarce. Limited data is a common issue in many physical systems in the real world, where the cost of acquiring data maybe prohibitively expensive. One effective strategy to overcome this challenge of data scarcity is to delibrately include prior knowlege as inductive biases in the model. Including structured knowledge about the data directly within the architecture can help steer the learning algorithm towards accurate solutions that generalize better \cite{raissi2019physics,battaglia2018relational,zhang2022gfinns,gruber2024efficiently}.

There have been many recent developments in learning Hamiltonian systems from data by leveraging mathematical properties of Hamiltonian systems in the form of inductive biases \cite{greydanus2019hamiltonian, bertalan2019learning, sanchez2019hamiltonian, chen2019symplectic}. Many of these works, including the popular Hamiltonian Neural Network \cite{greydanus2019hamiltonian} try to approximate the Hamiltonian of the system from the data, which can then be used to generate the phase flow of the system using numerical integrators. On the other hand, SympNets \cite{jin2020sympnets} directly learn the phase flow of the Hamiltonian systems by incorporating the symplectic structure, a mathematical property of Hamiltonian systems, into the neural network design.

\begin{figure}[!h]
    \centering
    \includegraphics[scale=0.45]{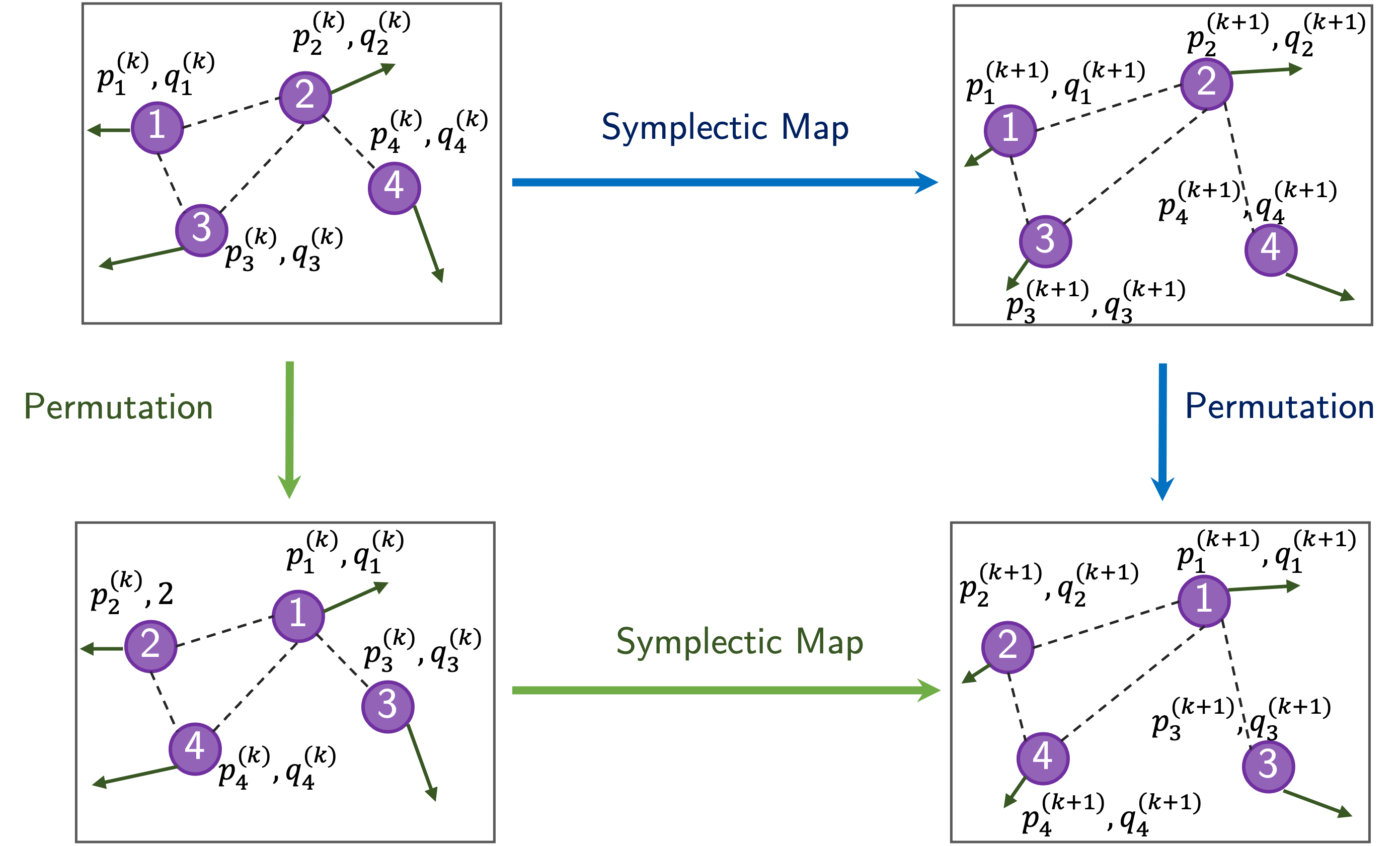}
    \caption{\textbf{Commutative diagram for a symplectically permutation equivariant map.} The SympGNN update from timestep $k$ to $k+1$ followed by a permutation, yields the same result as a permutation followed by SympGNN update, demonstrating that the map is equivariant to permutations.}
    \label{fig:commutative-diagram}
\end{figure}

Although these networks have exhibited potential in scientific applications, their implementation in complex, high-dimensional many-body systems continues to pose significant challenges. It was empirically observed that SympNets, along with other Hamiltonian-based neural networks including HNNs, struggle to identify high-dimensional many-body systems from data if no additional structures are added to the neural network \cite{tong2020symplectic}. One particular structure, that could potentially alleviate this issue is permutation equivariance, a property possessed by graph neural networks \cite{hamilton2020graph, bronstein2021geometric} (see Fig. \ref{fig:commutative-diagram}).

Graph neural networks have recently become prominent in learning physical many-body/multiparticle systems due to their ability to capture interactions between particles \cite{sanchez2020learning, gruber2024reversible, satorras2021n}. Graph neural networks are permutation equivariant networks that operate on graph structured data \cite{satorras2021n}. Herein, we combine the advantages of graph neural networks and and SympNets, and we propose \textbf{Symp}lectic \textbf{G}raph \textbf{N}eural \textbf{N}etworks (\textbf{SympGNN}). SympGNNs are symplecticlally permutation equivariant graph neural networks that can handle the task of high-dimensional system identification as well as node classification. 

The paper is organized as follows. In Section~\ref{sec:background}, we present the background material and relevant definitions. In Section~\ref{sec:sympgnn_motivation}, we introduce the motivation to design permutation equivariant SympNets, i.e., SympGNNs. In Section~\ref{sec:sympgnn}, we introduce the architecture of two graph neural network models, G-SympGNNs and LA-SympGNNs. In Section~\ref{sec:sympgnn_numerical}, we present three numerical simulations to demonstrate the capability of SympGNN in handling a variety of simulation tasks.

\section{Background}\label{sec:background}
This section provides an overview of the relevant material and terminology related to permutation equivariance and symplectic maps. These concepts will help in building symplectic graph neural networks (SympGNNs) that are permutationally equivariant and in understanding their structure.

Permutation invariance refers to the property where the output of a function remains unchanged regardless of the order of its inputs. In the following definitions, we will consider functions acting on $\bx \in \R^{n \times d}$, which represents a matrix of $n$ objects/entities, each with $d$-dimensional feature. 
\begin{definition}
    A map $f:\R^{n\times d} \to \R$ is called permutation invariant if $f(P\bx) = f(\bx)$ for all $\bx\in\R^{n\times d}$ for any permutation matrix $P$.
\end{definition}

Permutation equivariance is another important form of underlying structure in the data for several problems \cite{battaglia2018relational}. This refers to the outputs of a function maintaining the same order as the inputs, preserving the relational structures between the objects. For example, a function that predicts the positions and momentums of $n$ objects at a later timestep would be desired to be permutation equivariant.
\begin{definition}
    A map $\phi:\R^{n\times d} \to \R^{n\times d}$ is called permutation equivariant if $\phi(P\bx) = P\phi(\bx)$ for all $\bx\in\R^{n\times d}$ for any permutation matrix $P$.
\end{definition}

We now present the definition of a symplectic map. Symplectic maps are closely related to Hamiltonian systems in that the solution operator or the phase flow of a Hamiltonian system is symplectic \cite{hairer2006geometric}. The prior work on SympNets built composable modules that are symplectic to approximate the solution operator of Hamiltonian systems.
\begin{definition}
    A differentiable map $\phi: \R^{2d} \to \R^{2d}$ is called symplectic if the Jacobian matrix $\nabla \phi$ satisifies:
    \begin{equation*}
        \nabla \phi^T(\bx) J \nabla \phi(\bx) = J \quad \forall \bx \in \R^{2d},
    \end{equation*}
    where J is a matrix with $2d$ rows and $2d$ columns defined as:
    \begin{equation*}
        J:= \begin{pmatrix}
    0 & I_d \\ -I_d & 0
    \end{pmatrix},
    \end{equation*}
    and $I_d$ is the identity matrix with $d$ rows and $d$ columns.
\end{definition}

In this paper, we propose Symplectic Graph Neural Networks (SympGNN) that are both symplectic and permutation equivariant.
\begin{definition}
    Suppose $\bx = \begin{pmatrix}
        x_1 & \cdots & x_m
    \end{pmatrix}\in\R^{n\times m}$. Define the flatten map $fl :\R^{n\times m}\to \R^{nm}$ by $fl(\bx) = \begin{pmatrix}
        x_1\\
        \vdots \\
        x_m
    \end{pmatrix}$. A map $\phi:\R^{n\times 2d} \to \R^{n\times 2d}$ is called symplectically permutation equivariant if (i) $fl^{-1}\circ \phi \circ fl$ is symplectic and (ii) $\phi\left(\begin{pmatrix}
        P\bp & P\bq
    \end{pmatrix}\right) = \begin{pmatrix}
        P\Tilde{\bp} & P\Tilde{\bq}
    \end{pmatrix}$ for all $\bp, \bq\in\R^{n\times d}$ for any permutation matrix $P$ , where $\begin{pmatrix}
        \Tilde{\bp} & \Tilde{\bq}
    \end{pmatrix} = \phi\left(\begin{pmatrix}
        \bp & \bq
    \end{pmatrix}\right)$.
\end{definition}

\section{Motivation} \label{sec:sympgnn_motivation}
Consider a separable Hamiltonian system with $H(p,q) = T(p) + V(q)$:
\begin{equation}
\left\{ \begin{aligned}
  & \frac{dp(s)}{dt} =  - \nabla V(q(s)), \quad \forall s\in [0, \infty)\\
  & \frac{dq(s)}{dt} =  \nabla T(p(s)), \quad \forall s\in [0, \infty)\\
  & p(0)=p^{(0)}, \  q(0)=q^{(0)}  \ .
\end{aligned} \right.
\end{equation}
Here, the Hamiltonian $H$ typically represents the total energy of the system, $T$ represents the kinetic energy and $V$ represents the potential energy of the system. This equation describes the time evolution of the momentum $p(s)$ and positions $q(s)$ of the system over time $s$.

We denote the solution to the above equation at time $t$ by $p(t;p^{(0)}, q^{(0)})$ and $q(t;p^{(0)}, q^{(0)})$, to highlight the dependence on initial conditions. The solution operator $\phi_t$ is defined by 
\begin{equation}
    \phi_t\begin{pmatrix}
        p^{(0)}\\ q^{(0)}
    \end{pmatrix} := \begin{pmatrix}
        p(t;p^{(0)}, q^{(0)}) \\ q(t;p^{(0)}, q^{(0)})
    \end{pmatrix}.
\end{equation}

A general way to approximate $\phi_t$ would be to use the splitting scheme which takes the form \begin{equation}\label{eq:splitting}
    \varphi_t
    \begin{pmatrix}
    p^{(0)}\\q^{(0)}
    \end{pmatrix} := 
    \prod_{i=1}^l\left(
    \begin{pmatrix}
    I & 0 \\ k_i\nabla T_i & I
    \end{pmatrix}\begin{pmatrix}
    I & -h_i\nabla V_i \\ 0 & I
    \end{pmatrix}\right)\begin{pmatrix}
    p^{(0)}\\q^{(0)}
    \end{pmatrix},
\end{equation}
where $\sum_{i=1}^l h_i = \sum_{i=1}^l k_i = t$, $\sum_{i=1}^l V_i = V$, $\sum_{i=1}^l T_i = T$. 

The solution operator of the Hamiltonian system $\phi_t$ is proven to be a symplectic map. Our goal is to design parametric maps that preserve the symplectic structure to approximate this solution operator. One of the simplest family of symplectic maps from $\R^{2d}$ to $\R^{2d}$ is the following:
\begin{equation}\label{eq:2dmap}
    f_{up} \begin{pmatrix}
    p \\
    q
    \end{pmatrix} =
    \begin{bmatrix}
    I & \nabla U \\
    0 & I
    \end{bmatrix}
    \begin{pmatrix}
    p \\
    q
    \end{pmatrix},
    \quad
    f_{low} \begin{pmatrix}
    p \\
    q
    \end{pmatrix} =
    \begin{bmatrix}
    I & 0 \\
    \nabla U & I
    \end{bmatrix}
    \begin{pmatrix}
    p \\
    q
    \end{pmatrix}.
\end{equation}

It can be seen that \eqref{eq:2dmap} matches the form of \eqref{eq:splitting} if we replace $\nabla U$s by $-h_i\nabla V_i$ and $k_i\nabla T_i$. Therefore, we can interpret SympNet as being automatically learning a Hamiltonian system together with its splitting scheme, when the underlying Hamiltonian is \textit{separable}.

\begin{theorem}\label{symplecticpermutationequivariant}
    Suppose $V_i:\R^{n\times d} \to \R$ and $T_i:\R^{n\times d} \to \R$ are permutation invariant for all $i = 1, \cdots, l$, then the alternating composition of the following two parameterized functions:
 \begin{equation}\label{eq:sympgnn}
    \begin{split}
    &\mathcal{E}^{low}_i\begin{pmatrix} \bp & \bq \end{pmatrix}=\begin{pmatrix} \bp&\bq+\nabla T_i(\bp) \end{pmatrix} \quad \forall \bp,\bq\in\R^{n\times d},\\ &\mathcal{E}^{up}_i\begin{pmatrix} \bp & \bq \end{pmatrix}=\begin{pmatrix} \bp - \nabla V_i(\bq)&\bq \end{pmatrix} \quad \forall \bp,\bq\in\R^{n\times d},  \\
    &\varphi = \mathcal{E}^{up}_l\circ \mathcal{E}^{low}_l\cdots \mathcal{E}^{up}_1\circ \mathcal{E}^{low}_1 \quad \text{or} \quad \varphi= \mathcal{E}^{low}_l\circ \mathcal{E}^{up}_l\cdots \mathcal{E}^{low}_1\circ \mathcal{E}^{up}_1,
    \end{split}
    \end{equation}
    are symplectically permutation equivariant.
\end{theorem}

This inspires us to design $T_i$ and $V_i$ as neural networks with the permutation invariant property, in order to respect the permutation equivariance of $\varphi$. Theorem \ref{symplecticpermutationequivariant} serves as the starting point for the two SympGNN architectures presented in Section \ref{sec:sympgnn}.

\begin{remark}
    While SympNets are capable of learning a splitting scheme for separable Hamiltonian systems, they also possess the ability to learn systems with nonseparable Hamiltonians.
\end{remark}

\begin{remark}
    We will use the notation \begin{equation}
        \bx = \begin{pmatrix}
            x_1 & \cdots & x_m
        \end{pmatrix} = \begin{pmatrix}
            x^1 \\ \vdots \\ x^n
        \end{pmatrix}
    \end{equation} to represent the columnwise and rowwise expansion of any $\bx\in\R^{n\times m}$. Here, $x_i$ (index as a subscript) represents a column vector and $x^i$ (index as a superscript) represents a row vector.
\end{remark}

\section{Symplectic Graph Neural Networks (SympGNN)}\label{sec:sympgnn}
Consider an undirected graph $\mathcal{G} = (\mathcal{V}, E)$ consisting of $n$ nodes. Suppose the input node features are $\bx \in \R^{n\times m}$, where $m$ is the number of features for each node. The adjacency matrix of the graph is given by $A\in \R^{n\times n}$ such that $A_{ij} = A_{ji}, \forall i, j\in\{1,\cdots,n\}$, which could be either weighted or unweighted. A graph encoder is a function $\psi_{en}:\R^{n\times m} \to \R^{n\times 2d}$, such that 
\begin{equation}
    \psi_{en}\begin{pmatrix}
        x^1\\ \vdots \\ x^n
    \end{pmatrix} := \begin{pmatrix}
        \phi_{en}(x^1)\\ \vdots \\ \phi_{en}(x^n)
    \end{pmatrix} = \begin{pmatrix}
        p^1 &q^1\\ \vdots & \vdots \\ p^n & q^n
    \end{pmatrix},
\end{equation}
where $\phi_{en}:\R^{m}\to\R^{2d}$ can be (i) an identity map or (ii) a learnable affine map or (iii) a fully connected neural network. Similarly, we define a graph decoder to be $\psi_{de}:\R^{n\times 2d} \to \R^{n\times m_{out}}$, such that 
\begin{equation}
    \psi_{de}\begin{pmatrix}
        z^1\\ \vdots \\ z^n
    \end{pmatrix} := \begin{pmatrix}
        \phi_{de}(z^1)\\ \vdots \\ \phi_{de}(z^n)
    \end{pmatrix},
\end{equation}
for $\phi_{de}:\R^{2d}\to\R^{m_{out}}$.
A symplectic graph network (SympGNN) is defined by $\varphi_{SG} = \psi_{de}\circ \psi_{symp}\circ \psi_{en}$, where $\psi_{symp}$ is a symplectically permutation equivariant function. When choosing an affine map or a fully connected neural network as the encoder and decoder, the symplectic dynamics is preserved in the latent space. For physical systems, where a symplectic map between the input and output is desired, we can set the encoder and decoder to be identity maps. Additionally, in physical systems, the input feature vector $x^j = (p^j \quad q^j)$ can be a concatenation of the momentum $p^j$ and position $q^j$ of node $j$.

We now present two formulations of the SympGNN: i) G-SympGNN and ii) LA-SympGNN. Both formulations are permutation equivariant symplectic maps, and can be constructed in the following ways:

\begin{definition}[G-SympGNN]
    Let 
    \begin{equation}
        T_i^{(G)}(\bp):=\sum_{j=1}^n\phi_v^i(p^j)
    \end{equation}
    for any $\bp = \begin{pmatrix}
        p^1 \\\vdots\\p^n
    \end{pmatrix}\in\R^{n\times d}$, where $\phi_v^i:\R^{ d}\to \R$ is a node-level function which can be parameterized by fully-connected neural networks. Let 
    \begin{equation}
        - V_i^{(G)}(\bq; A):=
    \begin{dcases}
        \sum_{j=1}^n\sum_{(j,k)\in E}\phi_e^i(q^j, q^k, A_{jk}) \quad \text{if } A \ \text{is weighted}\\
        \sum_{j=1}^n\sum_{(j,k)\in E}\phi_e^i(q^j, q^k) \quad\text{otherwise}
    \end{dcases}
    \end{equation}
    for any $\bq = \begin{pmatrix}
        q^1 \\\vdots\\q^n
    \end{pmatrix}\in\R^{n\times d}$,  $\phi_e^i:\R^{2d(+1)}\to \R$ is an edge-level function which can be parameterized by fully-connected neural networks. The alternated composition of $\mathcal{E}_i^{up}, \mathcal{E}_i^{low}$ (see eq.~\ref{eq:sympgnn}) constructed with $T_i$, $V_i$ defined as $T_i^{(G)}$ and $V_i^{(G)}$, $i=1,\cdots,l$, are denoted as $\psi_{G}$. The G-SympGNN is defined by $\varphi_{GSG}:= \psi_{en}\circ\psi_{G}\circ\psi_{de}$.
\end{definition}

 \begin{theorem}
    $T_i^{(G)}$ and $V_i^{(G)}$ are permutation invariant for all $i\in\{1,\cdots, l\}$, so $\psi_{G}$ is symplectically permutation equivariant. Therefore, $\varphi_{GSG}$ is permutation equivariant.
\end{theorem}

In this formulation, we have chosen the kinetic energy function $T_i^{(G)}$ to be a parameteric function that maps the node-level momentum to the kinetic energy of each node and then sums these values across all nodes. Similarly, the potential energy function $V_i^{(G)}$ is a parameteric function that maps position (and edge weights) to potential energy for each edge and sums these values across all the edges. This design choice is motivated by our understanding of physical systems, where the kinetic energy of a system is derived from the momentum of individual particles, and the potential energy arises from the interaction between particles depending on their positions.

\begin{figure}[!h]
    \centering
    \includegraphics[scale=0.5]{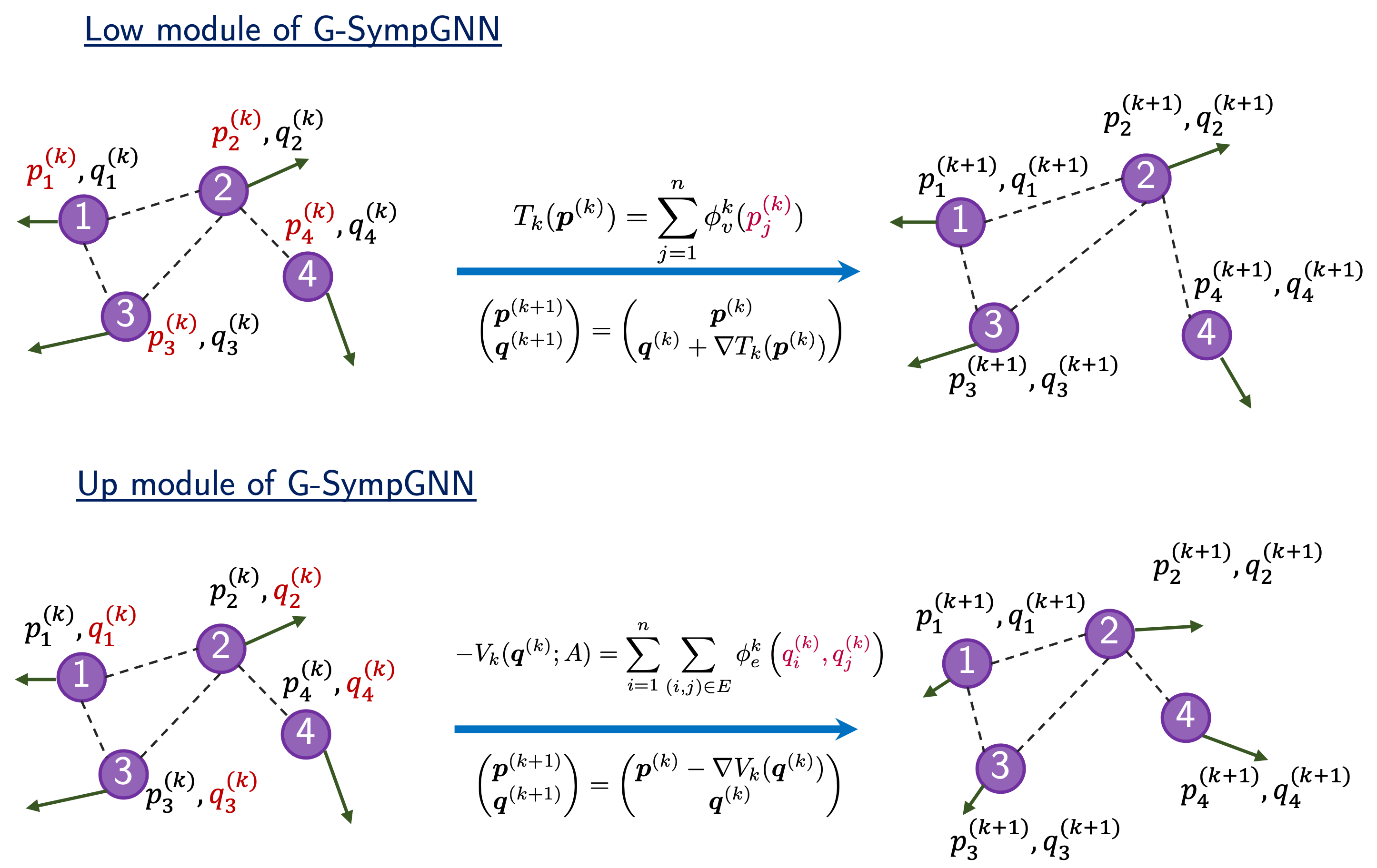}
    \caption{\textbf{Illustration of the low and up modules in G-SympGNN.} This figure shows the update rules from timestep $k$ to $k+1$ for the low module, which updates $\bq$, and the up module, which updates $\bp$. The G-SympGNN is constructed by alternating these low and up modules.}
    \label{fig:gsympgnn}
\end{figure}

Figure \ref{fig:gsympgnn} shows a visual representation of the update rules in the up module $\mathcal{E}_i^{up}$ and the low module $\mathcal{E}_i^{low}$ for G-SympGNN. The up module updates momentum $\bp$, while the low module updates the position $\bq$. The G-SympGNN involves the computation of the gradients of functions parameterized as neural networks and can be computationally expensive. Our second formulation, LA-SympGNN, avoids the need for gradient computation by using linear algebra operations for updates.

We now present the formulation of LA-SympGNN. The LA-SympGNN consists of linear modules ($L$) and nonlinear activation modules ($N$) as follows:
\begin{definition}[LA-SympGNN]
    Define $T_{i}^{(L)}, V_{i}^{(L)}, T_{i}^{(N)}, V_{i}^{(N)}:\R^{n\times d} \to \R$ as follows:
 \begin{equation}\label{eq:la_sympgnn}
    \begin{split}
    &T_{i}^{(L)}(\bp)=fl(\bp)^\top (K_i\otimes \square)fl(\bp),\\ -&V_{i}^{(L)}(\bq)=fl(\bq)^\top  (S_i\otimes \square)fl(\bq),\\
    &T_{i}^{(N)}(\bp)=\bm{1}_{1\times n}\left((\int\sigma)(\bp)\right)a_i,\\ -&V_{i}^{(N)}(\bq)=\bm{1}_{1\times n}\left((\int\sigma)(\bq)\right)b_i,
    \end{split}
    \end{equation}
    where $\otimes$ represent the Kronecker product of two tensors. $\square \in \R^{n\times n}$ represents a one-step linear graph message passing operator, which for instance can be (i) the degree normalized adjacency matrix, $\Tilde{A} = D^{-\frac{1}{2}}AD^{-\frac{1}{2}}$ or (ii) the graph Laplacian matrix, $L = D - A$. The learnable parameters are $S_{i}\in\R^{d\times d}$, $K_{i}\in\R^{d\times d}$, $a_{i}\in\R^{d\times 1}$ and $b_{i}\in\R^{d\times 1}$. The alternated composition of $\mathcal{E}_{i}^{up}, \mathcal{E}_{i}^{low}$ constructed with $T_{i}^{(L)}$, $V_{i}^{(L)}$, $T_i^{(N)}$, $V_i^{(N)}$, $i = 1, \cdots l$, are denoted as $\psi_{LA}$. The LA-SympGNN is defined by $\varphi_{LASG}:= \psi_{de}\circ\psi_{LA}\circ\psi_{en}$.
\end{definition}

\begin{theorem}
    $T_i^{(L)}$,  $V_i^{(L)}$, $T_i^{(N)}$ and $V_i^{(N)}$ are permutation invariant for all $i\in\{1\cdots l\}$, so $\psi_{LA}$ is symplectically permutation equivariant. Therefore, $\varphi_{LASG}$ is permutation equivariant.
\end{theorem}

The motivation behind this specific parameterizations of the kinetic energy $T_i$ and potential energy $V_i$ is to have linear algebra operations for updating $\bp$ and $\bq$, without requiring the computation of gradients. The LA-SympGNN can also be viewed as an extension of LA-SympNet (\cite{jin2020sympnets}) by exploiting the graph structure and imparting permutation equivariance as detailed in the appendix. Substituting the forms of $T_i$ and $V_i$ in eq. \ref{eq:sympgnn} and analytically computing the gradients yields the update rules for low and up modules.

The update rules for the linear low and up modules for the LA-SympGNN can be written as follows:
\begin{equation}\label{eq:linearlasympgnn}
    \begin{split}
    &\mathcal{E}^{low}_i\begin{pmatrix} \bp & \bq \end{pmatrix}=\begin{pmatrix} \bp&\bq+ \square \bp K_i'\end{pmatrix} \quad \forall \bp,\bq\in\R^{n\times d},\\ &\mathcal{E}^{up}_i\begin{pmatrix} \bp & \bq \end{pmatrix}=\begin{pmatrix} \bp+ \square \bq S_i' &\bq \end{pmatrix} \quad \forall \bp,\bq\in\R^{n\times d}.
    \end{split}
\end{equation}
Here, $K_i' = K_i + K_i^T$ and $S_i' = S_i + S_i^T$ are symmetric, and the square matrix $\square \in \R^{n \times n}$ can be any message passing operator such as $\tilde{A}$ or $L = D - A$.

The update rules for the activation low and up modules in LA-SympGNN can be written as:
\begin{equation}\label{eq:activationlasympgnn}
    \begin{split}
    &\mathcal{E}^{low}_i\begin{pmatrix} \bp & \bq \end{pmatrix}=\begin{pmatrix} \bp&\bq+ a_i^T \odot \sigma ( \bp )\end{pmatrix} \quad \forall \bp,\bq\in\R^{n\times d},\\ &\mathcal{E}^{up}_i\begin{pmatrix} \bp & \bq \end{pmatrix}=\begin{pmatrix} \bp+ b_i^T \odot \sigma( \bq )&\bq \end{pmatrix} \quad \forall \bp,\bq\in\R^{n\times d}.
    \end{split}
\end{equation}

Here, $\sigma(.)$ is any nonlinear activation function applied element-wise, and $\odot$ represents element-wise product. Note that $a_i^T \in \R^{1\times d}$ and $\sigma(\bp) \in \R^{n \times d}$ do not have compatible shapes for element-wise multiplication. We first broadcast $a_i^T$ to get a rectangular matrix of compatible shape $(\R^{n \times d})$. Similarly, broadcasting is done for $b_i^T$. A visual illustration of the linear low and activation low update modules for LA-SympGNN at a node-level is shown in Fig \ref{fig:la_sympgnn_low}. We have included the illustration for the linear up and activation up modules in the appendix.

\begin{figure}[!h]
    \centering
    \includegraphics[scale=0.5]{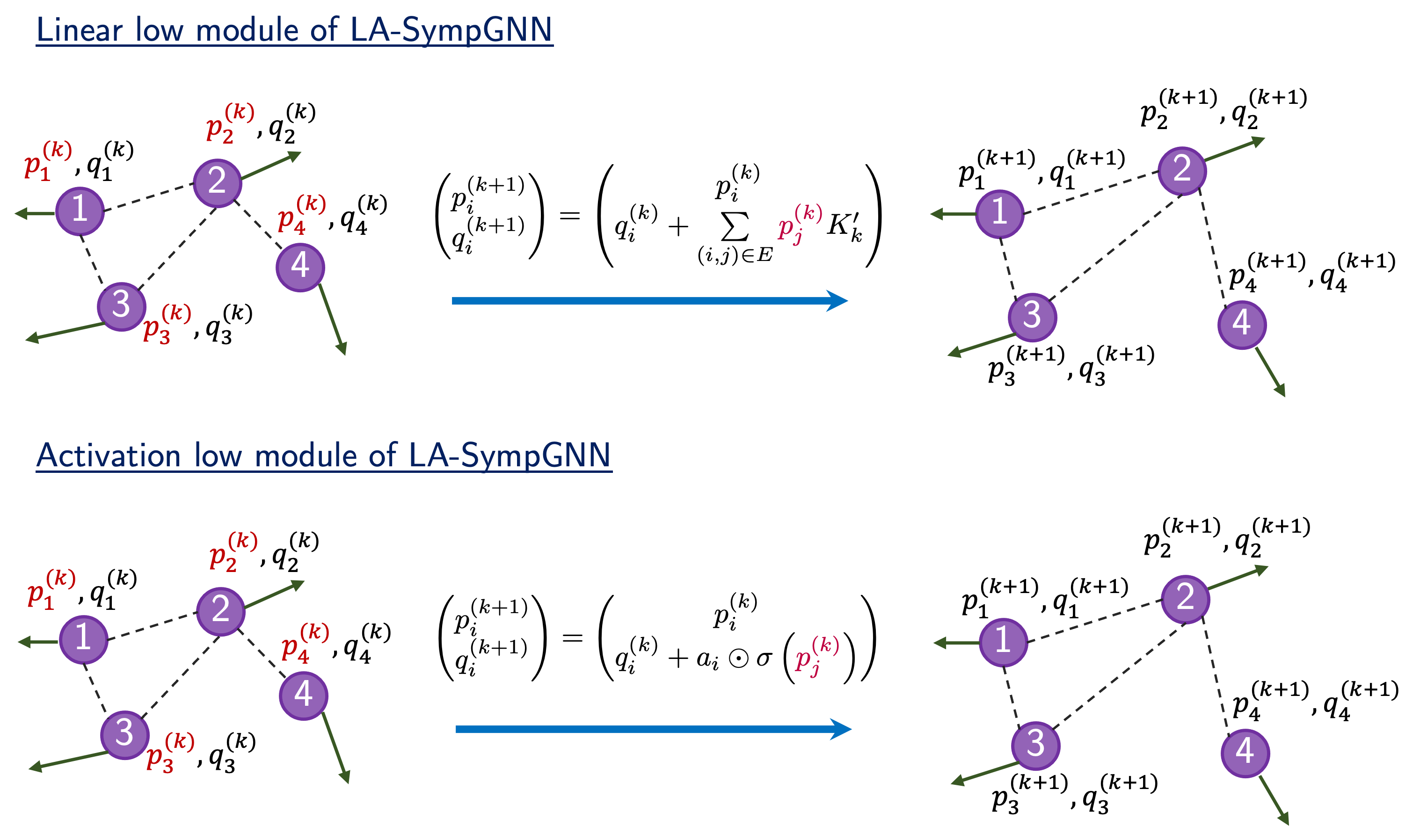}
    \caption{\textbf{Illustration of the linear low layer and the activation low layer in LA-SympGNN} for $\square = A$. The linear up layer updates $\bq$ based on its neigbhors through message passing.}
    \label{fig:la_sympgnn_low}
\end{figure}

\begin{remark}[Connection with GCN \cite{kipf2016semi} and GRAFF \cite{di2022understanding}]
    The update rule of a \textit{linear} Graph Convolution Network (GCN) at the $k$-th layer is 
    \begin{equation}
        \bx^{(k)} = \Tilde{A}\bx^{(k-1)}W_k,
    \end{equation}
    and the update rule of a \textit{linear} gradient-flow based graph neural network (GRAFF) at the $k$-th layer is 
    \begin{equation}
        \bx^{(k)} = \bx^{(k-1)} + \Tilde{A}\bx^{(k-1)}W_k - \bx^{(k-1)}diag(\omega_k) - \beta \bx^{(0)},
    \end{equation}
    where $W\in\R^{d\times d}$ is enforced to be symmetric. If we set $\beta = 0, \omega_k = 0$, then the update rules become
    \begin{equation}
        \bx^{(k)} = \bx^{(k-1)} + \Tilde{A}\bx^{(k-1)}W_k,
    \end{equation}
    which is basically \textit{linear} GCN with residual connection.
    If we set $\square = \Tilde{A}$ and consider only the linear layers in LA-SympGNN, then our update rule becomes similar to the update rules above in the sense that we are mapping 
    \begin{equation}
        \begin{split}
             &\bp^{(k)} = \bp^{(k-1)} + \Tilde{A}\bq^{(k-1)}S_k' \\
             &\bq^{(k)} = \bq^{(k-1)} + \Tilde{A}\bp^{(k)}K_k'. 
        \end{split}
    \end{equation}
    We also enforce $S_k'$, $K_k'$ to be symmetric. The difference lies in the fact that the dynamics of GRAFF follows a gradient flow while the dynamics of LA-SympGNN follows a symplectic flow.
\end{remark}

\section{Simulation results using SympGNN}\label{sec:sympgnn_numerical}
We consider two different types of tasks, system identification and node classification using SympGNN. The first two numerical simulations, coupled harmonic oscillator and the Lennard Jones particle examples are system identification tasks. The last example on the Cora dataset is an example of node classification.

\begin{figure}[!h]
    \centering
    \includegraphics[scale = 0.50]{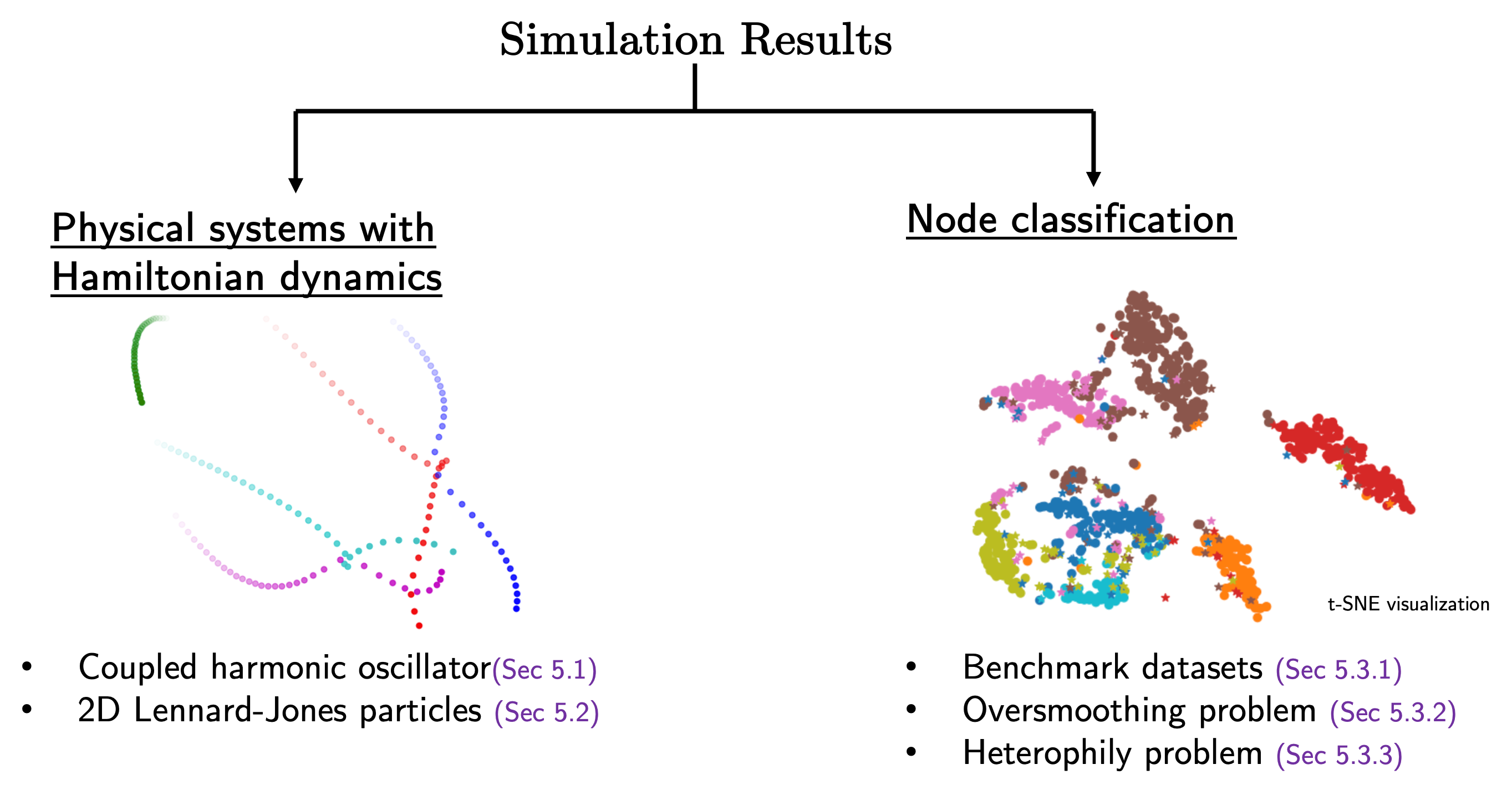}
    \caption{\textbf{Illustrative overview of the simulation results}. We present results on system identification in two physical systems with Hamiltonian dynamics: i) coupled harmonic oscillator and ii) 2D Lennard-Jones particles. We also present results on node classification in i) four benchmark datasets, ii) the oversmoothing problem and iii) the heterophily problem.}
    \label{fig:enter-label}
\end{figure}

For the system identification task, we are provided with the feature vector $\bx = (\bp, \bq) \in \R^{n\times 2d}$ directly, and the encoder / decoder will be set as identity map. The goal is to learn the dynamics from the history of $\{\bx(ht)\}_{t=1}^{T}$ and then predict the evolution of $\{\bx(ht)\}_{t=T+1}^{\Tilde{T}}$, where $h$ is the timestep between two consecutive observations. Specifically, in the training stage, we try to identify the "one-step-forward" solution map, i.e. $\varphi:\R^{n\times2d}\to\R^{n\times2d}$ such that $\varphi (\bx(ht)) =\bx(h(t+1)) \forall t\in\{1,\cdots,T-1\}$. Then in the test stage, we apply $\varphi$ iteratively to rollout the prediction $\varphi(\bx(hT)), \varphi\circ\varphi(\bx(hT)), \cdots, \varphi^{(\Tilde{T} - T)}(\bx(hT))$ to predict $\{\bx(ht)\}_{t=T+1}^{\Tilde{T}}$. Finally, the mean squared error on the rolled-out trajectory is calculated.

For the node classification task, we are provided with the feature vector $\bx = \begin{pmatrix}
    \bx_{train} , \bx_{val} , \bx_{test}
\end{pmatrix}^T\in \R^{(n_{1} + n_{2} + n_{3})\times m}$ and the classification labels $\by = \begin{pmatrix}
    \by_{train}, \by_{val}, \by_{test}
\end{pmatrix}^T \in \{1,\cdots,c\}^{(n_1 + n_2 + n_3)\times 1}$, where $n_1$ is the number of nodes in the training set, $n_2$ is the number of nodes in the validation set, $n_3$ is the number of nodes in the test set, and $c$ is the number of classes.
During the training stage, we aim to identify a map $\varphi:\R^{(n_1 + n_2 + n_3)\times m}\to \{1,\cdots,c\}^{(n_1 + n_2 + n_3)\times 1}$ such that $[\varphi(\bx)]_{1:n_1} \approx \by_{train}$.
Here, $[\varphi(\bx)]_{i:j}$ represents the $i$-th to $j$-th rows of $\varphi(\bx)$.
Then, the goal in the test stage is to use $[\varphi(\bx)]_{n_1 + n_2+1:n_1+n_2 + n_3}$ to predict $\by_{test}$ and evaluate the classification accuracy.
To summarize, during training we can observe the feature vector of all nodes and the entire graph structure, but the labels are available only of the training nodes.
This setting of node classification is referred to as transductive node classification.
The overall architecture for the node classification task can be written as follows:
\begin{align*}
    \bp(0) &= \psi_{en}^p(\bx) \quad \quad \quad \quad \text{(feature encoding for momentum)}, \\
    \bq(0) &= \psi_{en}^q(\bx) \quad \quad \quad \quad \text{(feature encoding for position)},\\
    \bp(T), \bq(T) &= \psi_{symp}(\bp(0), \bq(0)) \quad \quad \text{(symplectic layers)}, \\
    \hat{\by} &= \psi_{de} (\bq(T)) \quad \quad \quad \quad \quad \text{(output layer)}.
\end{align*}

\subsection{Coupled harmonic oscillator}
\begin{figure}[!htbp]
\includegraphics[width=\textwidth]{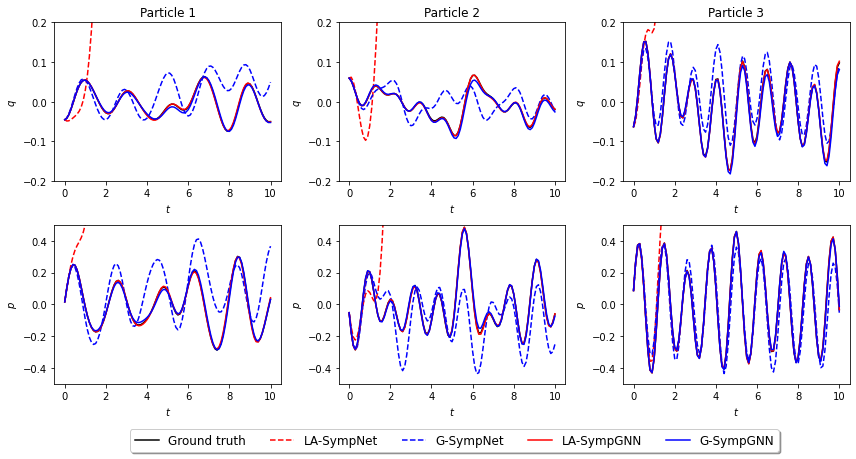}
\caption{\textbf{Comparison of SympNet vs SympGNN when the training set size $T = 500$.} We randomly sample three particles from the chain and plot the trajectory in the test window. SympGNN outperforms SympNet in matching the ground truth trajectory when provided limited training data. This is because more inductive biases (permutation equivariance) is embedded.}
\label{fig:spring}
\end{figure}
We consider a non-dimensionalized chain with $n=40$ particles connected by 39 springs with free boundary. All the particles share the same mass $m = 1$. The spring coefficient of the $i$-th spring, denoted as $k_i$, is sampled from the normal distribution $\mathcal{N}(5, 1.25^2)$, truncated below from $1$. The Hamiltonian of the system is given by 
\begin{equation*}
    H(\bp,\bq) = \frac{1}{2}\left(\sum_{i=1}^{n}p_i^2 + \sum_{i=1}^{n-1} k_i (q_i - q_{i+1})^2\right).
\end{equation*}
We further assume that all the particles are at rest at $t = 0$. The initial position of particles are generated from $U(-2.5, 2.5)$. We set the timestep between two observations $h = 0.1$. The number of test data points is fixed $\Tilde{T} - T = 100$. We further vary $T$ to check how the prediction MSE changes with the training sample size.

In this example, the weighted adjacency matrix $A$ is set to satisfy
\begin{equation}
    A_{ij} = \begin{dcases}
        0 \quad \text{if } |i - j| > 1 \ \text{or } \ i = j, \\
        k_j \quad \text{if } i - j = 1,  \\
        k_i \quad \text{if } j - i = 1.
    \end{dcases}
\end{equation}
We use the graph Laplacian $\square = L = D - A$ in the implementation of LA-SympGNN. 

\begin{figure}[!htbp]
\includegraphics[width=\textwidth]{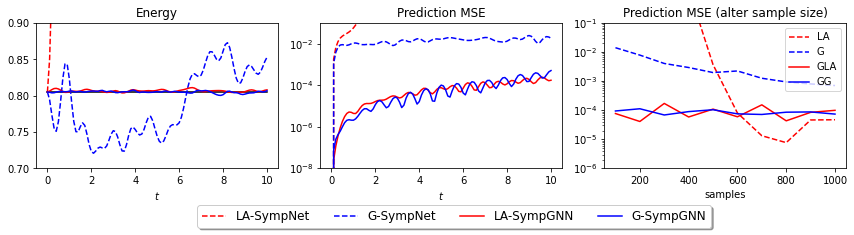}
\caption{\textbf{(Left and middle) Total energy and MSE of the prediction when $T = 500$. (Right) Prediction MSE as a function of training sample size $T$.} We can see that SympGNNs both conserves the energy better and produces lower MSE compared to SympNets when $T = 500$. However, when $T$ becomes larger, SympNets become comparable, or even better (G-SympNet) than SympGNNs, because they have more expressive power.}
\label{fig:spring_energy}
\end{figure}

We run the simulation with $300000$ iterations for LA-SympNet, G-SympNet, LA-SympGNN and G-SympGNN, first with the number of training samples $T = 500$. From Fig.~\ref{fig:spring}, we can see that SympGNN outperforms SympNet in matching the ground truth trajectory when provided with limited training data. This is because more inductive biases (permutation equivariance) is embedded. In Fig.~\ref{fig:spring_energy}, we plot the total energy and MSE of the prediction when $T = 500$ and the prediction MSE as a function of training sample size $T$. We can see that SympGNNs both conserves the energy better and produces lower MSE compared to SympNets when $T = 500$. However, when $T$ becomes larger, SympNets become comparable, or even better (G-SympNet) than SympGNNs, because they have more expressive power.

\subsection{2D molecular dynamics simulation with Lennard-Jones particles}\label{sec:2d_lj}
We consider $n = 2000$ Argon particles governed by the Lennard-Jones potential. The data is generated with the NVE ensemble so no thermostat is involved. The training and testing data are obtained by solving the Hamiltonian system with $H$ given by 
\begin{equation*}
    H(\bp,\bq) = \frac{1}{2}\sum_{i=1}^{n} \frac{1}{m_i} ||p_i||^2 + \sum_{i=2}^{n-1}\sum_{(i,j)\in E} V_{i,j}\Big(||q_i-q_j||\Big).
\end{equation*}
We construct a graph $\mathcal{G} = (\mathcal{V}, E)$ with $n$ nodes representing $n$ particles. An edge is constructed when the distance between two particles become less than a threshold $r_c$, i.e.
\begin{equation}\label{eq:neighbour_lj}
    A_{ij} = \begin{dcases}
        0 \quad \text{if } ||q_i - q_j|| > r_c, \ \text{or} \ i =j \\
        1 \quad \text{otherwise }. \\
    \end{dcases}
\end{equation} 
In classical molecular dynamics the Lennard-Jones potential is given by 
\begin{equation}
    V_{i,j}(r) = 4\epsilon\left((\frac{\sigma}{r})^{12}-(\frac{\sigma_{i,j}}{r})^{6}\right),
\end{equation} where $\epsilon, \sigma$ are constants representing the dispersion energy and the distance where the potential equals 0 respectively. Here we set the cutoff radius $r_c = 1.2$ nm. The simulation is conducted in a 2d periodic box with side length set to be $20$ nm. We use the Composite Stormer Verlet integrator (4th order) with $h = 0.01$ ps to generate the trajectory, then sample every 10 steps as our data for training and prediction. The training set consists of the data during the first 10 ps, and our goal is to predict the evolution of the system in the next 10 ps. 
\begin{figure}[!htbp]
  \centering
  \includegraphics[width=0.9\textwidth]{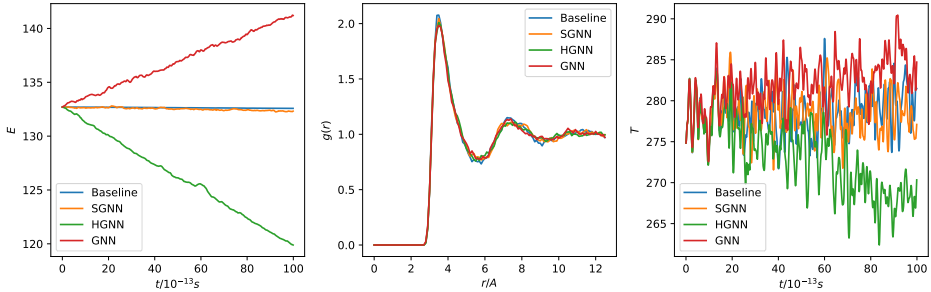}
  \caption{\textbf{Statistical quantities of the predicted trajectories by MPNN, HGNN and G-SympGNN.} It can be seen that G-SympGNN conserves the energy better than MPNN and HGNN. The left plot shows the evolution of per-particle energy (in $k_B[J]$, where $k_B$ is the Bolzmann constant) with respect to time, the center plot shows the variation of radial distribution function $g(r)$ with distance $r$, and the right plot shows evolution of temperature (in $K$) with time.}
  \label{fig:lj_2d}
\end{figure}
We compared the performance of G-SympGNN with the message passing neural network (GNN, \cite{gilmer2020message}) and Hamiltonian graph neural network (HGNN, \cite{sanchez2019hamiltonian}). It can be seen that G-SympGNN (denoted as SGNN in the legend) conserves the energy better than GNN and HGNN as shown in Fig~\ref{fig:lj_2d}.

\subsection{Performance on real world node-classification tasks}

\subsubsection{Benchmark datasets}
In this example, we evaluate the performance of LA-SympGNN on node classification task using four real-world benchmark datasets that exhibit diverse characteristics. Details of the datasets are presented in Table \ref{tab:benchmark_statistics}.We use the same train, validation and test splits taken from \cite{pei2020geom}, to ensure consistency and fair comparison of our results with prior works. The performance metrics reported are the mean accuracy and standard deviation on the test data across the 10 train/validation/test splits. The accuracies for the baseline metrics we compare against are taken from \cite{choi2023gread}.

In our experiments we utilized four variants of the SympGNN. The first variant is the LA-SGNN. The second variant is identical to the first, except it omits the use of activation layers. The third and fourth variants are modifications of the second model by including additional batch normalization and dropout layers. More details about these four variants and the code can be found in our GitHub repository: \url{https://github.com/alanjohnvarghese/LA-SympGNN}. To train our models, we used the Adam optimizer to minimize the cross-entropy loss, including a weight decay term for regularization. The hyperparameters in the model were tuned using Weights \& Biases \cite{wandb} using a standard random search with 500 counts, similar to the approach in \cite{choi2023gread}. The search space used for hyperparameter tuning is detailed in the appendix.

\begin{table}[ht]
\centering
\caption{Statistics and properties of the 4 benchmark datasets considered.}
\label{tab:benchmark_statistics}
\begin{tabular}{l r r r r }
\toprule \toprule
\textbf{Attribute} & \textbf{Film} & \textbf{Squirrel} & \textbf{Chameleon} & \textbf{Cora} \\ \midrule \midrule
Classes    & 5 & 5 & 5 & 6 \\ 
Features   & 932 & 2,089 & 235 & 1,433 \\ 
Nodes      & 7,600 & 5,201 & 2,277 & 2,708 \\ 
Edges      & 26,752 & 198,353 & 31,371 & 5,278 \\ 
Hom. Ratio & 0.22 & 0.22 & 0.23 & 0.81 \\
\bottomrule \bottomrule
\end{tabular}
\end{table}

\begin{table}[ht]
\centering
\caption{Node classification accuracy on benchmark datasets. Highlighted are the top \textcolor{red}{first}, \textcolor{Green}{second} and \textcolor{blue}{third} results.}
\label{tab:benchmark_results}
\begin{tabular}{@{}lcccc@{}}
\toprule \toprule
\textbf{Method} & \textbf{Film} & \textbf{Squirrel} & \textbf{Chameleon} & \textbf{Cora} \\ 
\midrule \midrule
Homophily & 0.22 & 0.22 & 0.23 & 0.81 \\
GCN       & $27.32 \pm 1.19$ & $53.43 \pm 2.01$ & $64.82 \pm 2.24$ & $86.98 \pm 1.27$ \\
GAT       & $27.44 \pm 0.89$ & $40.72 \pm 1.55$ & $60.26 \pm 2.50$ & $87.30 \pm 1.10$ \\
LINKX     & $36.10 \pm 1.55$ & \textcolor{Green}{$61.81 \pm 1.80$} & $68.42 \pm 1.38$ & $84.64 \pm 1.13$ \\
GRAND     & $35.62 \pm 1.01$ & $40.05 \pm 1.50$ & $54.67 \pm 2.54$ & $87.36 \pm 0.96$ \\
GRAFF (max)& $37.11 \pm 1.08$ & $59.01 \pm 1.31$ & \textcolor{Green}{$71.38 \pm 1.47$} & \textcolor{Green}{$88.01 \pm 1.03$} \\
GREAD (max)& \textcolor{red}{$37.90 \pm 1.17$} & {$59.22 \pm 1.44$} & \textcolor{red}{$71.38 \pm 1.31$} & \textcolor{red}{$88.57 \pm 0.66$} \\
\\
\hline\\
LA-SGNN & $35.22 \pm 0.87$ & $56.38 \pm 1.23$ & $70.11 \pm 1.16$ & \textcolor{blue}{$87.65 \pm 0.81$} \\
L-SGNN (i)& $35.24 \pm 1.16$ & \textcolor{blue}{$60.13 \pm 1.28$} & \textcolor{blue}{$71.10 \pm 2.07$} & $86.62 \pm 1.21$ \\
L-SGNN (ii)& \textcolor{Green}{$37.22 \pm 1.02$} & $59.52 \pm 1.61$ & $70.02 \pm 1.44$ & $86.12 \pm 0.95$ \\ 
L-SGNN (iii)& \textcolor{blue}{$37.11 \pm 1.06$} & \textcolor{red}{$62.56 \pm 0.97$} & $70.75 \pm 1.66$ & $86.18 \pm 1.14$ \\ 
\bottomrule \bottomrule
\end{tabular}
\end{table}

The outcomes of the numerical experiments using LA-SympGNN are summarized in Table \ref{tab:benchmark_results}. On the Squirrel dataset, our model achieves state-of-the-art accuracy. For the remaining three datasets, we are within the top three accuracies. However, we note that the node classification task for these benchmark datasets are very sensitive to hyperparameter tuning \cite{di2022understanding}. In the next few examples, we show the ability of our model to alleviate the oversmoothing problem and heterophily problem in node classification, which are more robust to the choice of hyperparameters and a better indicator of the goodness of model.

\subsubsection{Oversmoothing problem}
The oversmoothing problem presents a significant challenge to the performance of various graph neural network (GNN) architectures \cite{rusch2023survey,yan2022two}. This issue manifests itself as a decline in model performance with increasing depth of the network. Specifically, as the number of layers in these models increases, the node representations tend to converge towards indistinguishable vectors, thereby obscuring the distinctions between individual nodes. The phenomenon of oversmoothing arises due to the repeated application of message passing operations, wherein each node aggregates features from its neighbors \cite{chamberlain2021grand}.

In this numerical example, we utilize the Cora dataset \cite{mccallum2000automating} to evaluate the performance of SympGNN in addressing the oversmoothing problem, reporting the test accuracy for models with different depths. The Cora dataset consists of 2,708 scientific publications, each represented in a node, and classified into $c=7$ categories. These publications are represented as nodes in a citation graph, where an edge between two nodes signifies that one publication cited the other. The feature vector $\bx\in\R^{2708\times 1433}$ includes the content information of each publication, typically abstracts represented as bag-of-words feature vectors, and the citation relationships among them. We compare the performance of LA-SympGNN against different kinds of GCN architectures, which are designed specifically for the purpose of alleviating oversmoothing phenomena \cite{chamberlain2021grand,di2022understanding}:
\begin{itemize}
    \item GCN: $\bx^{(k)} = \sigma(\Bar{A}\bx^{(k-1)}W_k)$, where $\Bar{A} = D^{-\frac{1}{2}}(A+I)D^{-\frac{1}{2}}$.
    \item GCN with residual: $\bx^{(k)} = \bx^{(k-1)} + \sigma(\Bar{A}\bx^{(k-1)}W_k)$,
    \item symmetric GCN with residual: $\bx^{(k)} = \bx^{(k-1)} + \sigma(\Bar{A}\bx^{(k-1)}(W_k + W_k^\top))$,
    \item symmetric GCN with residual and no self-loop: $\bx^{(k)} = \bx^{(k-1)} + \sigma(\Tilde{A}\bx^{(k-1)}(W_k + W_k^\top))$
    \item GRAND: $\frac{\partial}{\partial t} \bx(t)= (\textbf{A}(\bx(t)) - \textbf{I}) \bx(t)$
\end{itemize}

In the LA-SympGNN model, we set $\square = \Tilde{A} = D^{-\frac{1}{2}}AD^{-\frac{1}{2}}$. Both the encoder ($\psi_{en}$) and decoder ($\psi_{de}$) are set as affine functions (linear layers). The dimension of feature embedding is set to $d = 64$. The LA-SympGNN model consisted of alternating upper linear \& activation and lower linear \& activation modules. We used a dropout layer with $p = 0.5$ after the decoder. To train the model we used the Adam optimizer with a learning rate of $3e-3$ and used weight decay of $3e-2$ for regularization. The train/validation/test splits follow the Planetoid split protocol \cite{yang2016revisiting}. 

\begin{figure}[!h]
    \centering
    \includegraphics[scale = 0.55]{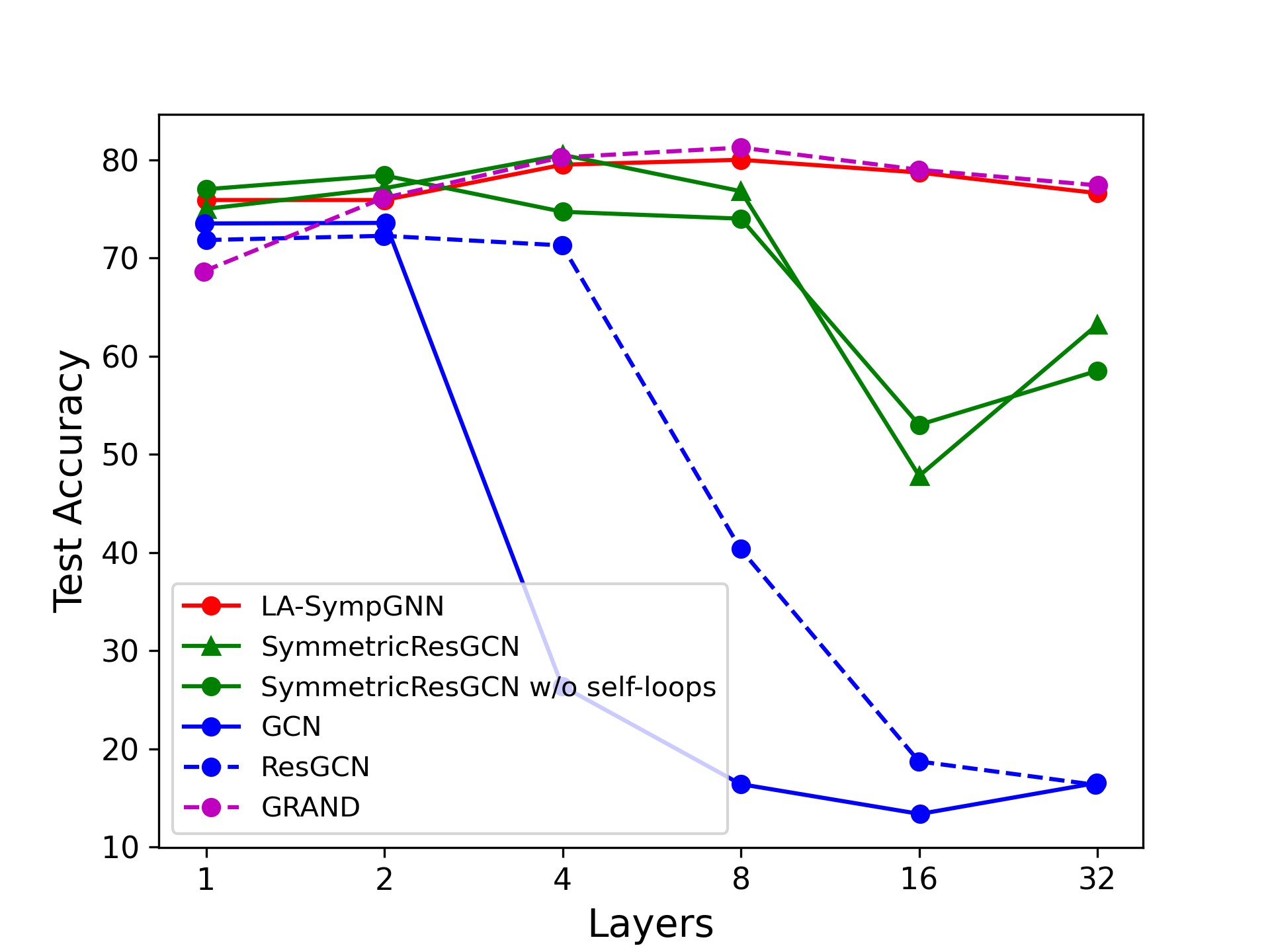}
    \caption{\textbf{The prediction accuracy of LA-SympGNN and GCN variants with respect to different number of layers, on the Cora dataset.} We observe that the accuracy of LA-SympGNN drops only marginally as the depth of the model increases. For the other models, except GRAND, we observe a significant drop in performance with depth due to oversmoothing.}
    \label{fig:oversmoothing}
\end{figure}

It can be seen from Fig.~\ref{fig:oversmoothing} that the LA-SympGNN avoids the oversmoothing problem when compared with the 4 GCN benchmarks. Only a slight drop in accuracy is observed for LA-SympGNN as the depth of the network increases, similar to GRAND.

\subsubsection{Heterophily problem}
In this example, we evaluate the performance of SympGNN in tackling the heterophily problem, a scenario where the performance of many GNN architectures degrades in heterophilic graphs. Heterophilic or low-homophilic graphs are characterized by a high tendency for neighboring nodes to belong to different classes. We assess the performance of SympGNN using the synthetic Cora \cite{zhu2020beyond} dataset. The synthetic Cora dataset shares the same features as Cora dataset, but the labels are manually generated by a modified perferential process for homophily ratios varying from 0 to 1 in steps of 0.1. The homophily ratio, which measures the fraction of edges connecting nodes with identical labels, is formally defined as \cite{ma2021homophily}:
\begin{equation}
    \mathcal{H} = \frac{1}{|E|} \sum_{(j,k) \in E} \mathbbm{1}\left( y_j  = y_k\right),
\end{equation}
where $|E|$ represents the number of edges, and $\mathbbm{1}(.)$ is the indicator function.

In the LA-SympGNN model, we set $\square = \Tilde{A} = D^{-\frac{1}{2}}AD^{-\frac{1}{2}}$. Both the encoder ($\psi_{en}$) and decoder ($\psi_{de}$) are set as affine functions (linear layers). The dimension of feature embedding is set to $d = 64$. We used a dropout layer with $p = 0.1$ after the decoder. To train the model, we used the Adam optimizer with a learning rate of $1e-3$ and a weight decay of $2e-1$ for heterophilic graphs ($\mathcal{H} \le 0.5$) and a weight decay of $5e-2$ for homophilic graphs ($\mathcal{H}> 0.5$). For each homophily ratio, we trained the model with the three different data splits of the synthetic Cora dataset and report the mean and standard deviation.


\begin{figure}[!h]
    \centering
    \includegraphics[scale = 0.55]{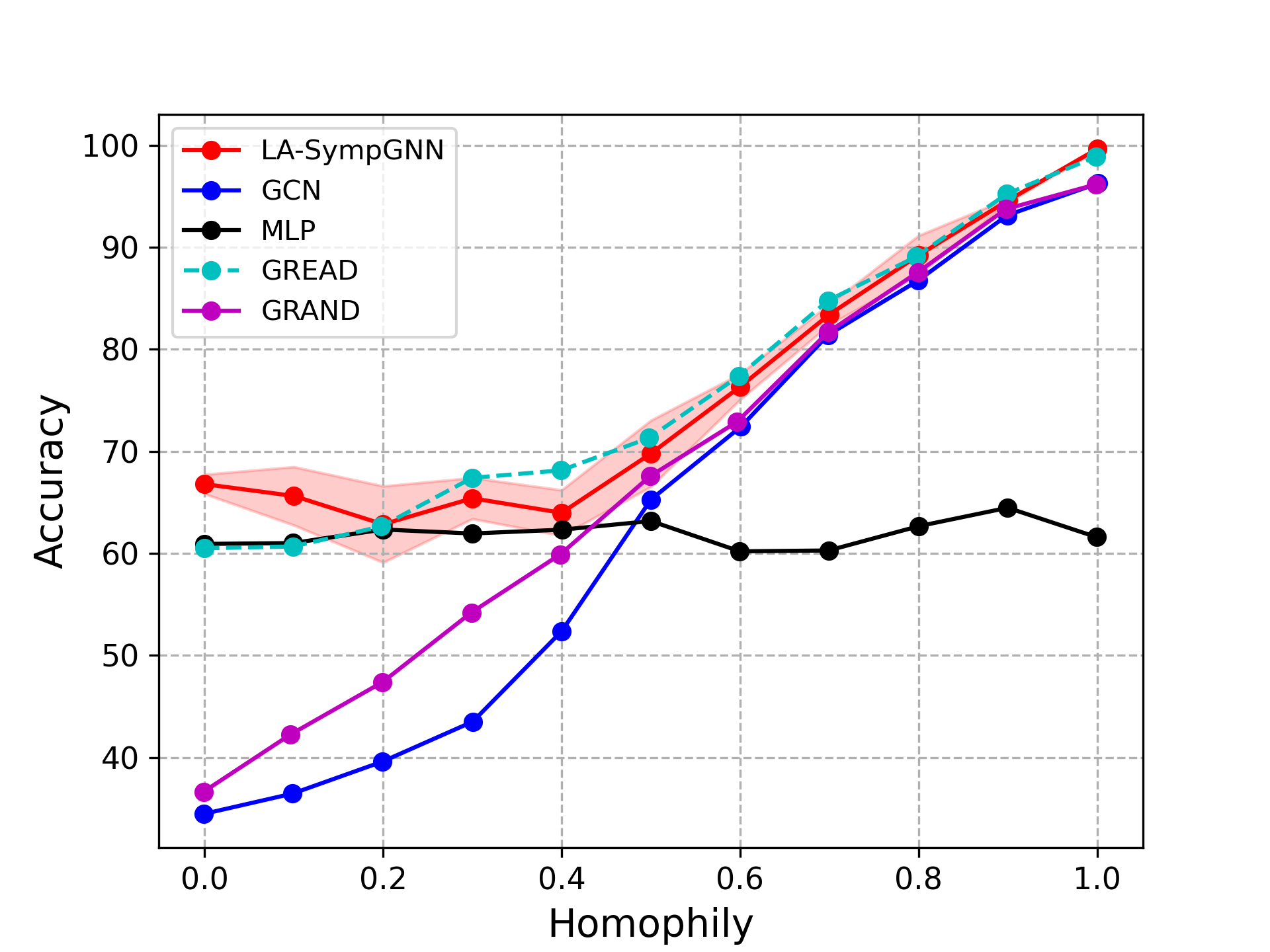}
    \caption{\textbf{The prediction accuracy of LA-SympGNN, GCN and MLP with respect to homophily ratio on the synthetic Cora dataset.} LA-SympGNN achieves MLP-like performance in the low-homophily regime and GCN-like performance in the high-homophily regime, effectively adapting to both types of graphs.}
    \label{fig:heterophily}
\end{figure}
In Fig. \ref{fig:heterophily}, we plot the prediction accuracy of LA-SympGNN at different homophily levels. The accuracy of GCN is low for heterophilic graphs and increases as the graph becomes homophilic. In the case of MLP although low, the prediction accuracy is independent of the homophily ratio, as the prediction does not depend on the feature vectors of the neighboring nodes due to the absence of message passing. We observed that LA-SympGNN achieves GCN-like performance on the high-homophily regime, and MLP-like performance on the low-homophily regime. The diffusion based GRAND also struggles to maintain accuracy in the low-homophily regime. The performance of our model is comparable to GREAD and is slightly better than GREAD for extremely low homophily ratios (0.0 and 0.1).


\section{Summary}
In this paper, we presented two new permutation equivariant symplectic networks: i) G-SympGNN and ii) LA-SympGNN. We demonstrated the effectiveness of the SympGNN on identification of Hamiltonian systems, specifically the coupled Harmonic oscillator,  and molecular dynamics of 2000 particles subject to a two-dimensional Lennard-Jones potential. We further showed the ability of LA-SympGNN to perform node classification, achieving accuracy comparable to the state-of-the-art. Finally, we demonstrated empirically the ability of SympGNN to alleviate the oversmoothing and heterophily problems. Possible future directions involve extending this framework to handle dissipative systems as well as time-dependent Hamiltonian systems.

\section*{Acknowledgements}
We would like to acknowledge support by the DOE SEA-CROGS project (DE-SC0023191) and the ONR Vannevar
Bush Faculty Fellowship (N00014-22-1-2795).
\bibliography{main.bib}

\newpage

\appendix
\section{Motivation behind the parameterizations for LA-SympGNN}
The motivation behind the parametrization of LA-SympGNN arises from extending our previous work on LA-SympNet. The LA-SympNet consists of linear modules:
\begin{equation*}
\ell_{up} \begin{pmatrix} p \\ q \end{pmatrix} =  \begin{bmatrix} I & S' \\ 0 & I \end{bmatrix} \begin{pmatrix} p \\ q \end{pmatrix} ,
\quad 
\ell_{low} \begin{pmatrix} p \\ q \end{pmatrix} = \begin{bmatrix} I & 0 \\ S' & I \end{bmatrix} \begin{pmatrix} p \\ q \end{pmatrix} ,
\end{equation*}
and activation modules:
\begin{equation*}
\mathcal{N}_{up} \begin{pmatrix} p \\ q \end{pmatrix} = \begin{bmatrix} I & \tilde{\sigma}_a \\ 0 & I \end{bmatrix} \begin{pmatrix} p \\ q \end{pmatrix}, 
\quad 
\mathcal{N}_{low} \begin{pmatrix} p \\ q \end{pmatrix} = \begin{bmatrix} I & 0 \\ \tilde{\sigma}_a & I \end{bmatrix} \begin{pmatrix} p \\ q \end{pmatrix},
\end{equation*}
where $S' \in \R^{d \times d}$ is symmetric, $\tilde{\sigma}_a(.) = diag(a)\sigma(.)$, and $p,q, a \in \R^d$.

Let us first focus on the linear modules, and extend them to exploit the graph structure in the data by using the adjacency matrix $A \in \R^{n \times n}$, for the simple case of $n$ particles and each particle having a one-dimensional momentum and position, therefore, the momentum and position of the system would be $p,q \in \R^n$. The graph structure can be included in the linear modules by parameterizing $S'$ as $S' \odot A$:
\begin{equation*}
\ell_{up} \begin{pmatrix} p \\ q \end{pmatrix} =  \begin{bmatrix} I & S' \odot A \\ 0 & I \end{bmatrix} \begin{pmatrix} p \\ q \end{pmatrix} ,
\quad 
\ell_{low} \begin{pmatrix} p \\ q \end{pmatrix} = \begin{bmatrix} I & 0 \\ S'\odot A & I \end{bmatrix} \begin{pmatrix} p \\ q \end{pmatrix},
\end{equation*}
where $S' \in \R^{n \times n}$ is again symmetric and $p,q \in \R^n$.

Now, we extend this to systems where the position and momentum of each particle is high dimensional, $\bp, \bq \in \R^{n \times d}$. In this case, we ensure permutation equivariance and include the graph structure in the parameterization by employing the Kronecker product $\otimes$ as follows:
\begin{equation*}
\ell_{up} \begin{pmatrix} fl(\bp) \\ fl(\bq) \end{pmatrix} =  \begin{bmatrix} I & S' \otimes A \\ 0 & I \end{bmatrix} \begin{pmatrix} fl(\bp) \\ fl(\bq) \end{pmatrix} ,
\quad 
\ell_{low} \begin{pmatrix} fl(\bp) \\ fl(\bq) \end{pmatrix} = \begin{bmatrix} I & 0 \\ S'\otimes A & I \end{bmatrix} \begin{pmatrix} fl(\bp) \\ fl(\bq) \end{pmatrix},
\end{equation*}
where $S \in \R^{d \times d}$ is symmetric, $A \in \R^{n \times n}$ and $fl(\bp), fl(\bq) \in \R^{nd}$. Note that $S\otimes A$ is also symmetric, and has shape $\R^{nd \times nd}$ compatible for multiplication with $fl(\bp)$ or $fl(\bq)$.

Now, recall the identity, $\left( B \otimes A\right) fl(X) = fl(AXB^T)$. Therefore, the up and low modules can be written as:
\begin{equation*}
\ell_{up} \begin{pmatrix} fl(\bp) \\ fl(\bq) \end{pmatrix} = \begin{pmatrix} fl(\bp) + fl(A\bq S') \\ fl(\bq) \end{pmatrix},
\quad 
\ell_{low} \begin{pmatrix} fl(\bp) \\ fl(\bq) \end{pmatrix} = \begin{pmatrix} fl(\bp) \\ fl(\bq) + fl(A\bp S') \end{pmatrix},
\end{equation*}

Now, motivated from Theorem \ref{symplecticpermutationequivariant}, we would like to find functions $V$ and $T$ such that $-\nabla_{\bq} V = A\bq S'$ and $ \nabla T_{\bp} = A \bp S'$.

The function $V$, that satisfies the above equation can be found by expressing $A\bq S'$ as  $ A\bq S' = \nabla_{\bq} \sum\limits_{i,j} A_{ij} \left< q_i, Sq_j\right>$, here $S' = S + S^T$. Therefore, $-V = \sum\limits_{i,j} A_{ij} \left< q_i, Sq_j\right>  =  fl(\bq)^\top  (S\otimes A)fl(\bq)$.

Similarly, we can find the function $T$ by expressing $A \bp S'$ as $ A\bp S' = \nabla_{\bp} \sum\limits_{i,j} A_{ij} \left< p_i, Kp_j\right>$, here $S' = K + K^T$, to distinguish between the parameters in the up and low modules. Therefore, $T = \sum\limits_{i,j} A_{ij} \left< p_i, Kp_j\right>  =  fl(\bp)^\top  (K\otimes A)fl(\bp)$.

Note that in this specific example we have used the adjacency matrix to incorporate the graph structure into our parameterization. In practice, we can use any one-step linear graph message passing operator (we denote using $\square$) instead of $A$.

Now, we will focus on the activation modules in the LA-SympNet, $\tilde{\sigma}_a(.) = diag(a) \sigma(.)$. Note that here, there is no mixing between the $d$ components of momentum and position. Each component of the momentum and position undergoes a nonlinear mapping and gets multiplied by a trainable scalar. In the case of systems with multiple particles and high dimensional momentum and position, we will share the scalar for each component across particles. Therefore, we will have $\tilde{\sigma}_a(.) = \sigma(.) diag(a)$, where $\sigma(.) \in \R^{n \times d}$ and $diag(a) \in \R^{d \times d}$.

We now want to find a function $T$, such that $\nabla_p T = \sigma(\bp) diag(a)$. This is given by $T = \bm{1}_{1\times n}\left((\int\sigma)(\bp)\right)a$. Similarly, we want to find the function $V$ such that $-\nabla_q V = \sigma(\bq) diag(b)$. Here, we have used $b$ to distinguish the parameters in the up and low modules. Now, we have $-V = \bm{1}_{1\times n}\left((\int\sigma)(\bq)\right)b$.

\begin{figure}[!h]
    \centering
    \includegraphics[scale=0.5]{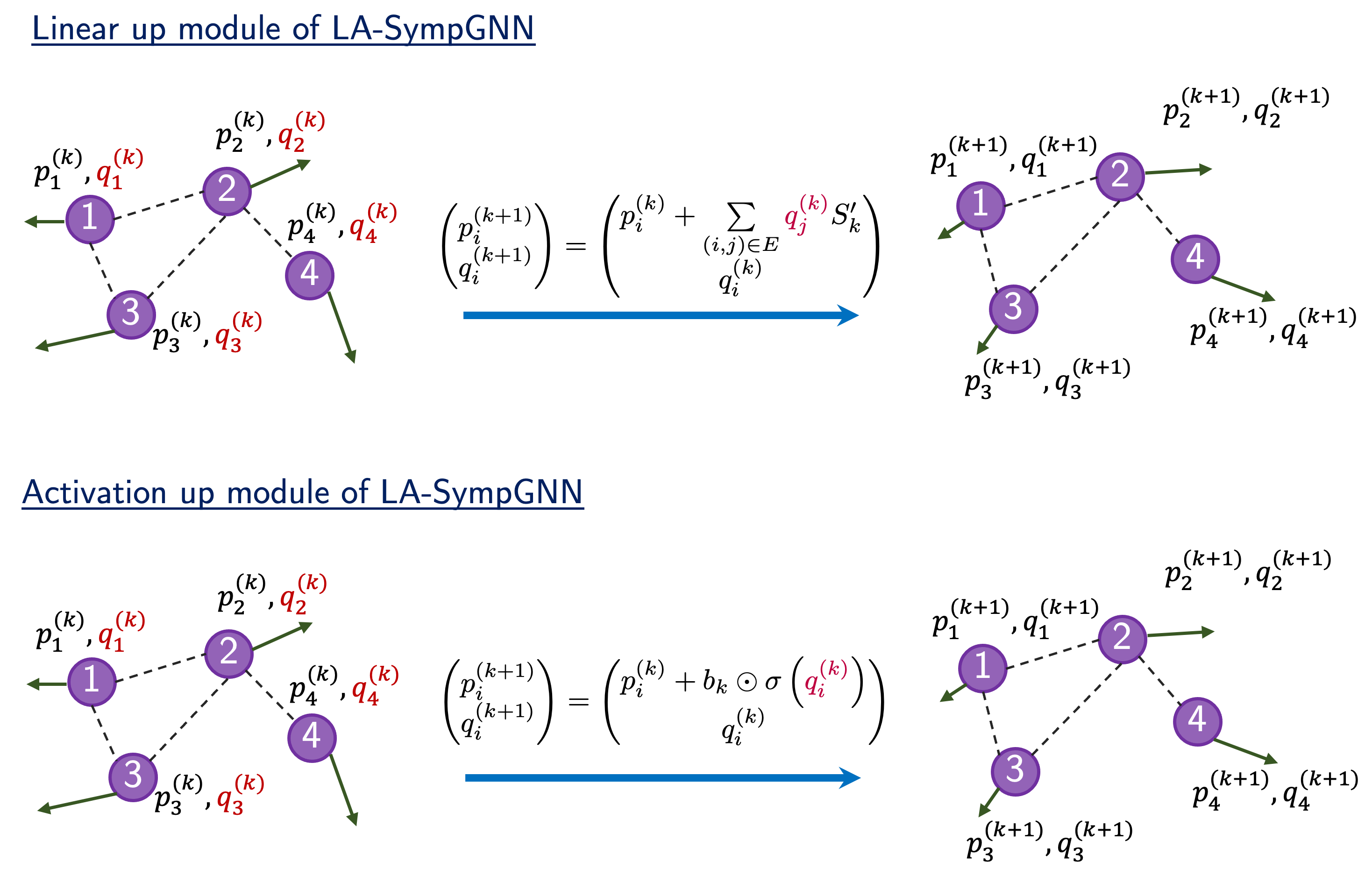}
    \caption{\textbf{Illustration of the linear up module and the activation up module in LA-SympGNN} for $\square = A$. The linear up layer updates $\bp$ based on its neigbhors through message passing.}
    \label{fig:la_sympgnn_up}
\end{figure}

The above parameterizations for potential and kinetic energy lead to linear algebraic updates for $\bp$ and $\bq$, without requiring the computation of gradients of $T$ and $V$. The updates for the up modules of linear and activation LA-SympGNN are illustrated in Fig. \ref{fig:la_sympgnn_up}.

\section{Node classification task on benchmark datasets}
\subsection{Hyperparameter tuning}
The search-space used for hyperparameter tuning for the real-world node classification task is given in Table \ref{tab:searchspace}. The use of batch normalization and the use of non-linearities in the encoder and decoder are also additional hyperparameters that are tuned for.

\begin{table}[!h]
    \centering
    \begin{tabular}{cc}
    \toprule
         \textbf{Hyperparameter}& \textbf{Search Space}\\
    \midrule 
         Epochs& 500\\
         Learning rate& \textit{log uniform}[5e-4, 1e-2]\\
         Weight decay& \textit{log uniform}[1e-6, 5e-2]\\
         Dropout& \textit{uniform}[0, 0.6] \\
         Input Dropout& \textit{uniform} [0, 0.6]\\
         Latent dimension& \{8,16,32,64,128\} \\
         Number of layers& \{1,2,3,4,5\}\\
    \bottomrule
    \end{tabular}
    \caption{Hyperparameter search-space used to fine tune our model for the node classification task on real-world experiments.}
    \label{tab:searchspace}
\end{table}
\end{document}